\documentclass{article}
\usepackage{spconf,amsmath,graphicx,hyperref}

\usepackage{cite}                  
\usepackage{amsmath,amssymb,amsfonts} 
\usepackage{algorithmic}            
\usepackage{graphicx}               
\usepackage{textcomp}               
\usepackage{xcolor}                 
\usepackage{caption}                
\usepackage{newtxtext,newtxmath}    
\usepackage{multirow}               
\usepackage{arydshln}               
\usepackage{comment}                
\usepackage{booktabs}

\title{InstanceRSR: Real-World Super-Resolution via Instance-Aware Representation Alignment}
\name{Zixin Guo$^{1}$, Kai Zhao$^{2}$, Luyan Zhang$^{3*}$\thanks{Corresponding author: zhang.luya@northeastern.edu}}
\address{%
$^{1}$Tongji University, $^{2}$Western Sydney University, $^{3}$Independent Researcher \\
}
\begin{document}
\ninept
\maketitle
\begin{abstract}
Existing real-world super-resolution (RSR) methods based on generative priors have achieved remarkable progress in producing high-quality and globally consistent reconstructions. However, they often struggle to recover fine-grained details of diverse object instances in complex real-world scenes. This limitation primarily arises because commonly adopted denoising losses (e.g., MSE) inherently favor global consistency while neglecting instance-level perception and restoration. To address this issue, we propose InstanceRSR, a novel RSR framework that jointly models semantic information and introduces instance-level feature alignment. Specifically, we employ low-resolution (LR) images as global consistency guidance while jointly modeling image data and semantic segmentation maps to enforce semantic relevance during sampling. Moreover, we design an instance representation learning module to align the diffusion latent space with the instance latent space, enabling instance-aware feature alignment, and further incorporate a scale alignment mechanism to enhance fine-grained perception and detail recovery. Benefiting from these designs, our approach not only generates photorealistic details but also preserves semantic consistency at the instance level. Extensive experiments on multiple real-world benchmarks demonstrate that InstanceRSR significantly outperforms existing methods in both quantitative metrics and visual quality, achieving new state-of-the-art (SOTA) performance.
\end{abstract}

\section{Introduction}
\label{sec:intro}

Deep learning based~\cite{dai2024periodicity, liu2024timebridge, liu2024wftnet, dai2024ddn} real-world super-resolution (RSR) must handle complex and unknown degradations that vary with imaging conditions. Such degradations often lead to local structural blurring or ambiguity, thereby limiting the fidelity of reconstruction. Traditional approaches typically employ multi-stage random degradation modeling to simulate blur, noise, and compression artifacts \cite{cai2019toward}. Although these methods have achieved some success, purely end-to-end modeling remains insufficient for ensuring high-quality perceptual reconstruction \cite{yang2025diffusion}.

\begin{figure}[t]
\centering
\includegraphics[width=0.95\columnwidth]{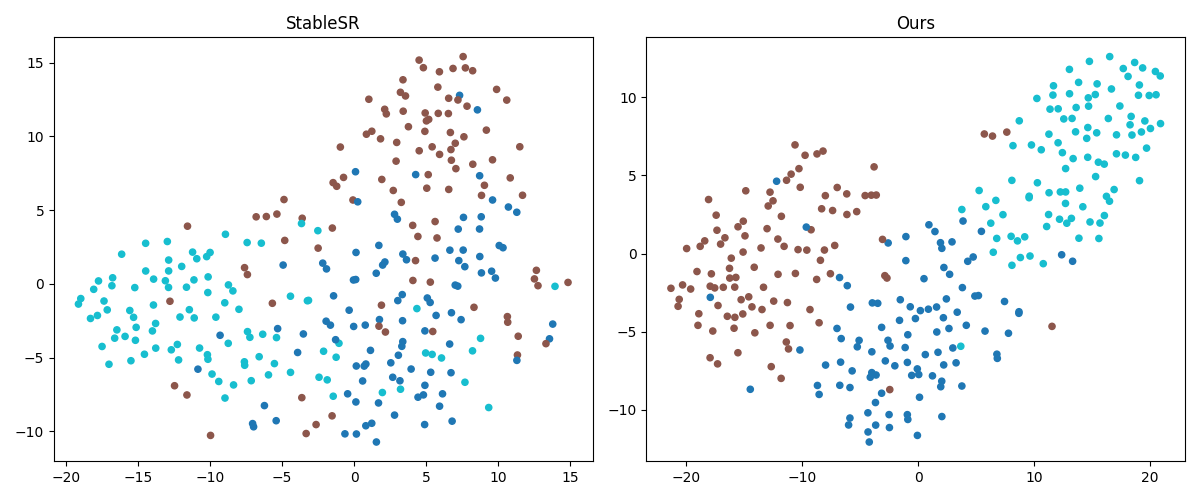} 
\caption{t-SNE visualization of intermediate feature representations comparing StableSR and our method. Each point corresponds to a sample, color-coded by its semantic category.}
\label{fig1}
\end{figure}

Recently, the emergence of diffusion models has significantly advanced perceptual quality and consistency in image generation. Existing RSR methods generally use low-resolution (LR) images as conditional inputs to preserve global semantics, while optimizing with denoising losses \cite{yang2025personalized}. These conditions can be introduced internally (similar to SR3 \cite{saharia2023image}) or externally via prior injection (e.g., StableSR \cite{wang2024exploiting}), which enforces the integration of global information and prior knowledge to improve perceptual reconstruction. However, denoising loss is inherently biased toward restoring local high-frequency semantics, with relatively weak constraints on low-frequency components. This limitation becomes particularly pronounced in multi-instance scenarios: when an image contains multiple objects, fine-grained textures and boundaries are more easily lost or misinterpreted under severe degradation, and existing methods struggle to effectively model and restore instance-level details.

\begin{figure*}[ht]
\centering
\includegraphics[width=2\columnwidth]{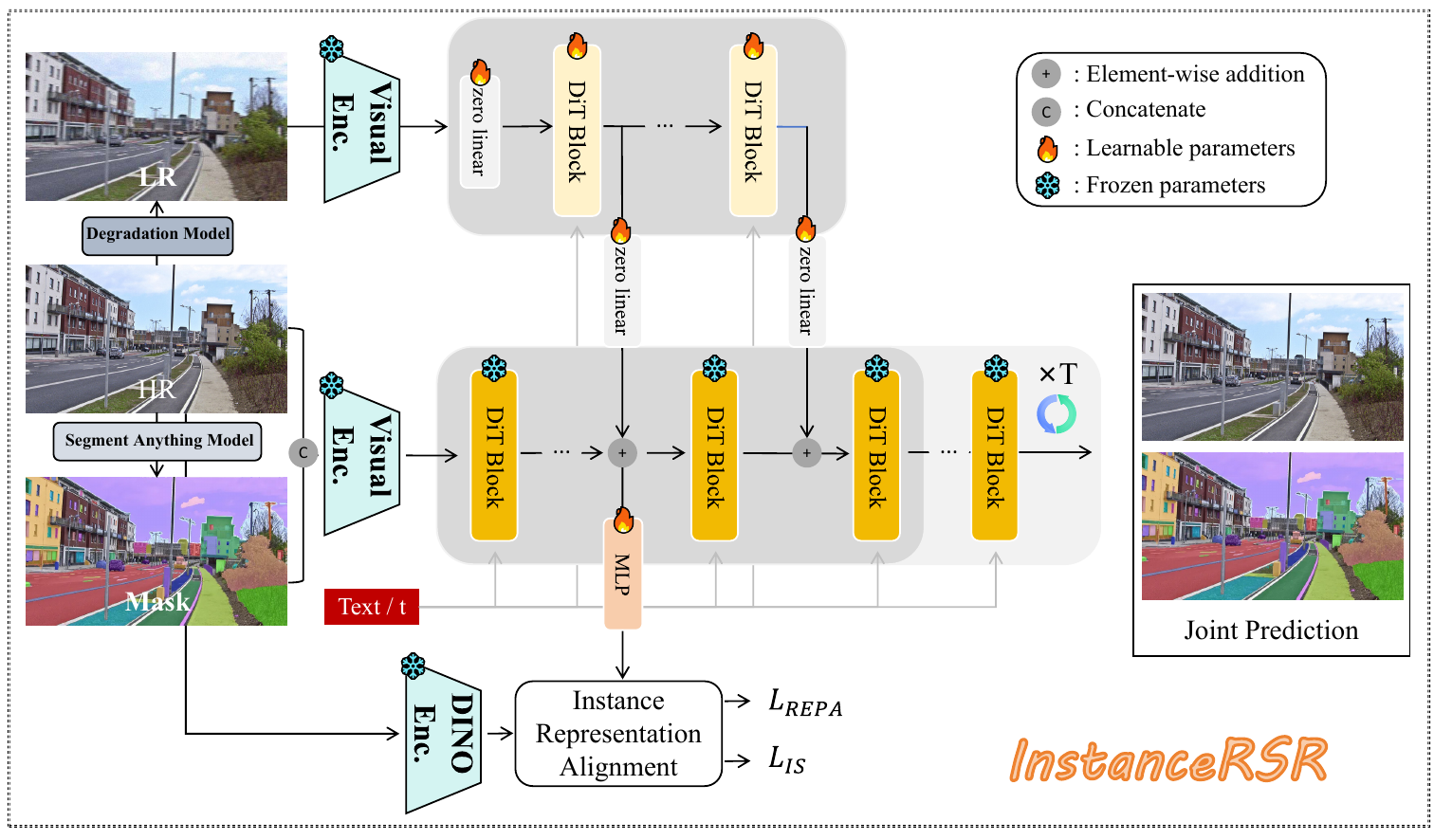} 
\caption{Overview of the proposed InstanceRSR framework. The model integrates instance masks and representation alignment into a DiT-based pipeline, where frozen visual encoders, backbone and semantic guidance jointly enhance instance awareness.}
\label{fig2}
\end{figure*}

Recent studies demonstrate that strengthening representation learning can substantially enhance the semantic perception ability of generative models. For example, REPA \cite{yu2024repa} aligns hidden representations of diffusion Transformers with features from pretrained visual encoders, enabling faster convergence and improved generation quality; Dispersive Loss \cite{wang2025diffuse} encourages feature dispersion in latent space, promoting stronger discriminability and disentanglement—similar to contrastive learning—thereby improving semantic feature separation. Nevertheless, current SR generative models still suffer from insufficient representations. As illustrated in Fig. \ref{fig1}, SOTA methods such as StableSR exhibit a certain degree of ``stickiness” in instance representations, making it difficult to sufficiently disentangle object features, which in turn restricts fine-grained detail restoration.

To address these issues, we propose InstanceRSR, a RSR framework that introduces an instance-aware representation alignment mechanism. The core idea is to establish joint alignment of semantic and instance features within the generator’s latent space. Specifically, we incorporate the LR image as a global semantic condition into a pretrained Diffusion Transformer (DiT) \cite{peebles2023scalable}, while jointly modeling image data and semantic segmentation maps to explicitly constrain instance-level information. Building on this, we align representations such that feature vectors of the same semantic category remain consistent with corresponding instance-level features, thereby constructing an instance-aware latent space. This mechanism is further enhanced with feature dispersion regularization, which guides the generative process to better differentiate and focus on the correct object details in latent space (see Fig. \ref{fig1}). Extensive experiments on multiple real-world benchmarks demonstrate that InstanceRSR achieves SOTA performance in both perceptual quality and reconstruction fidelity. In particular, the visual results highlight our method’s ability to produce sharper, more consistent, and semantically coherent reconstructions, especially in fine detail restoration.

\section{Method}
\label{sec:method}

\subsection{Overview of the InstanceRSR}


\begin{figure*}[th]
\centering
\includegraphics[width=2.0\columnwidth]{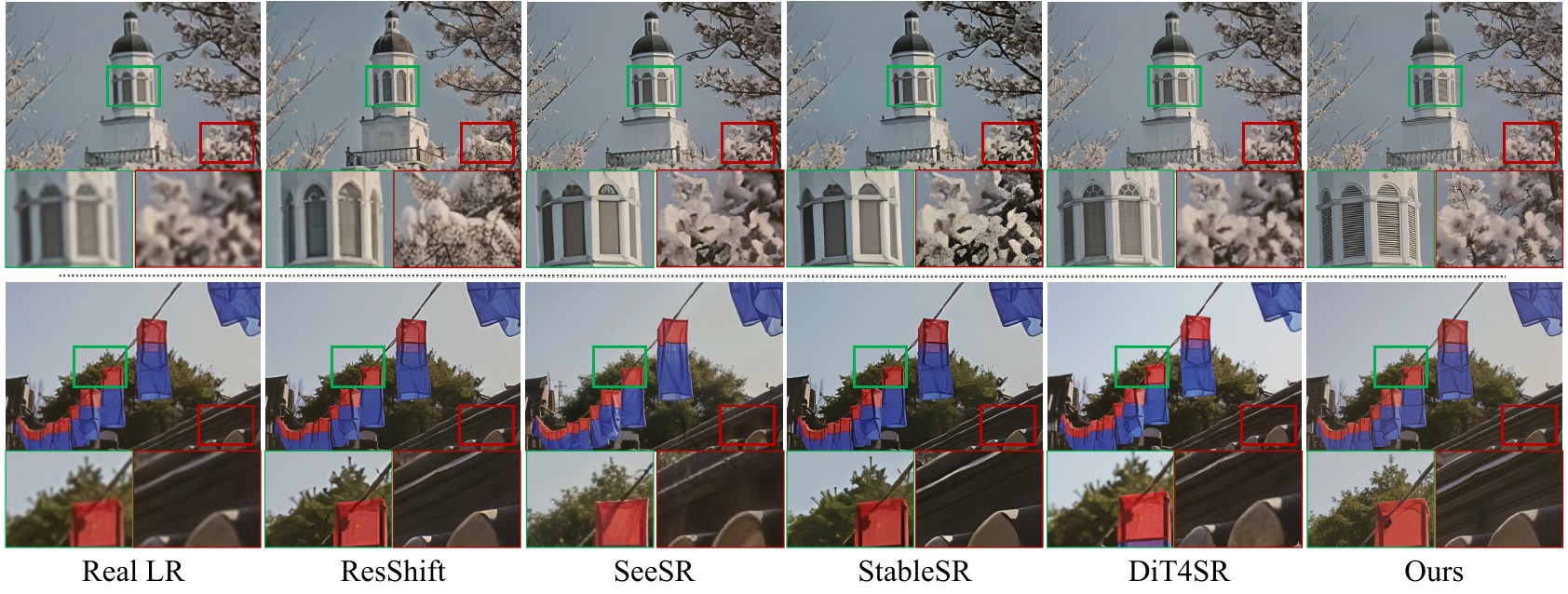} 
\caption{Visual comparison on the RealSR dataset. Competing methods tend to produce geometry distortions, over-smoothing, or noisy artifacts. In contrast, our InstanceRSR restores sharp structures and fine textures with clear boundaries and artifact-free details.}
\label{fig3}
\end{figure*}

As shown in Fig. \ref{fig2}, the proposed InstanceRSR framework is built upon a pretrained DiT. Given a real-world high-resolution (HR) image $\textbf{x}$, we first apply the Segment Anything Model (SAM) \cite{kirillov2023segment} to obtain its semantic segmentation map $\textbf{m}$. Then, a low-resolution (LR) image $\textbf{y}$ is synthesized via the real degradation model following Real-ESRGAN \cite{wang2021real}:

\begin{equation}
\textbf{y} = \big((\textbf{x} \otimes k) \downarrow_{s} + a\big) \downarrow_{s'} + j ,
\end{equation}
where $k$ denotes the blur kernel, $\otimes$ represents convolution, $\downarrow_{s}$ and $\downarrow_{s'}$ are downsampling operators with different scales, $a$ is additive noise, and $j$ denotes compression artifacts. This process generates degraded observations that better approximate real imaging conditions.  

In the encoding stage, $(\textbf{x, m, y})$ are independently mapped into the latent space by visual encoders, yielding $(\textbf{z}_x, \textbf{z}_m, \textbf{z}_y)$. To simultaneously model image content and semantic masks, we slightly expand the input and output dimensions of DiT (twice the original size). We denote the concatenated latent of the image and its semantic mask as \(\textbf{z} = [\textbf{z}_x; \textbf{z}_m]\). During training, we apply the standard forward noising process to this joint latent:

\begin{equation}
q(\textbf{z}_t \mid \textbf{z}_{t-1})=\mathcal{N}\big(\textbf{z}_t; \sqrt{\alpha_t}\, \textbf{z}_{t-1},\ (1-\alpha_t)\mathbf{I}\big),
\end{equation}
where \(\{\alpha_t\}\) is the predefined noise schedule and \(\bar{\alpha}_t=\prod_{s=1}^t \alpha_s\). The reverse (generative) process is parameterized by the DiT backbone augmented with the ControlNet branch \cite{zhang2023adding}, producing a conditional denoiser \(\epsilon_\theta(\textbf{z}_t, t \mid \mathcal{C})\) where \(\mathcal{C}\) denotes conditioning information derived from the degraded observation \(y\) (and timestep / text embeddings). Concretely, ControlNet injects shallow, zero-initialized blocks into corresponding backbone layers; the learned outputs of these blocks are linearly projected and added to the backbone features, yielding a conditional feature map \(\textbf{f}\) that modulates the denoising dynamics. The reverse step is therefore modeled as
\begin{equation}
p_\theta(\textbf{z}_{t-1}\mid \textbf{z}_t, \mathcal{C}) = \mathcal{N}\Big(\textbf{z}_{t-1};\ \mu_\theta(\textbf{z}_t,t,\mathcal{C},\textbf{f}),\ \Sigma_\theta(t)\Big),
\end{equation}
and implemented via the common epsilon-prediction parameterization used in diffusion models. At each diffusion timestep, $\textbf{f}$ is projected and aligned with features $\textbf{d}$ extracted from $x$ using a pretrained DINO encoder \cite{dino}, which strengthens instance-level representation learning. The overall training objective combines the standard denoising loss with the alignment term $\mathcal{L} = \mathcal{L}_{\text{denoise}} + \mathcal{L}_{\text{align}}$ . The denoising loss is

\begin{equation}
\mathcal{L}_{\text{denoise}} = \mathbb{E}_{t,z_0,\epsilon}\big\|\epsilon - \epsilon_\theta(\textbf{z}_t,t,\mathcal{C})\big\|_2^2.
\end{equation}

We will elaborate on the alignment loss mechanism in the next subsection. During inference, only the degraded image $\textbf{y}$ is used as the condition. After $T$ diffusion steps, the model jointly reconstructs the HR latent representation $\textbf{z}_x$ and its corresponding semantic representation $\textbf{z}_m$, thereby achieving instance-aware RSR after visual decoding.

\subsection{Instance-aware Representation Learning}

Although jointly modeling the semantic segmentation map can partially enhance the model's understanding of instance-level features, the denoising loss of the diffusion model essentially functions like positive-pair alignment in contrastive learning, lacking the repulsive effect of negative samples. As a result, the learned internal representations may remain ambiguous. To address this issue, we introduce an additional representation alignment supervision to eliminate ambiguity in the latent and improve their discriminative capability. Specifically, the HR image $\mathbf{x}$ is first encoded using a pretrained DINOv2 model \cite{dino} to obtain semantic features $\mathbf{d}$. Then, the hidden features $\mathbf{f}$ are projected via a learnable projection head $MLP_\phi$ into the same feature space as $\mathbf{d}$. The representation alignment loss is defined as:  
\begin{equation}
\mathcal{L}_{\mathrm{REPA}}(\theta,\phi) := -\mathbb{E}_{\mathbf{x},\epsilon}\left[\frac{1}{N}\sum_{n=1}^{N}\mathrm{sim}\bigl(\mathbf{d}[n],\,MLP_\phi(\mathbf{f}[n])\bigr)\right],
\end{equation}
where $n$ indexes the patches and $\mathrm{sim}(\cdot,\cdot)$ denotes cosine similarity. This loss maximizes the similarity between the projected hidden features of the diffusion model and the DINOv2 features at a per-patch level, encouraging the hidden features to learn noise-invariant and semantically rich representations. 

However, DINOv2 primarily captures semantic information such as object concepts and textures, and is less sensitive to instance-level spatial scale differences. Therefore, semantic alignment alone is insufficient to distinguish different instances within the same scene. To address this, we propose an instance-scale (IS) loss to provide additional instance-level supervision and enhance the model's global perspective. Specifically, the SAM is used to segment each object instance in the image and assign a unique ID. A random scale target $s_{\mathrm{target},i}$ is then assigned to each instance to prevent the model from learning trivial mappings. For the hidden feature $\mathbf{f}_{i,n}$ of the $n$-th patch of the $i$-th instance, we enforce its $\ell_2$ norm to converge to the corresponding scale target:
\begin{equation}
\mathcal{L}_{\mathrm{IDSA}} = \mathbb{E}_{i,p}\left[\left(\|\mathbf{f}_{i,n}\|_2 - s_{\mathrm{target},i}\right)^2\right].
\end{equation}
This loss encourages the learned representations to reflect the scale information of each instance, thereby enhancing the discriminability of different instances from a global perspective. Finally, the total alignment loss is defined as $\mathcal{L}_{\mathrm{align}} = \lambda_{\text{REPA}}\mathcal{L}_{\mathrm{REPA}} + \lambda_{\text{IS}}\mathcal{L}_{\mathrm{IS}}.$

\begin{table}[t]
\centering
\tiny
\setlength{\tabcolsep}{1.8pt} 
\renewcommand{\arraystretch}{1.4} 
\scalebox{1.05}{\begin{tabular}{c|c|cccccccc}
\hline
Datasets & Metrics & Real-ESRGAN & ResShift & StableSR & SeeSR & DiffBIR & OSEDiff & DiT4SR & Ours \\
\hline
\multirow{5}{*}{DrealSR} 
& LPIPS $\downarrow$ & 0.282 & 0.353 & 0.273 & 0.317 & 0.452 & 0.297 & 0.365 & \textbf{0.265} \\
& MUSIQ $\uparrow$ & 54.267 & 52.392 & 58.512 & 65.077 & 65.665 & 64.692 & 64.950 & \textbf{66.120} \\
& MANIQA $\uparrow$ & 0.490 & 0.476 & 0.559 & 0.605 & 0.629 & 0.590 & 0.627 & \textbf{0.635} \\
& ClipIQA $\uparrow$ & 0.409 & 0.379 & 0.438 & 0.543 & 0.572 & 0.519 & 0.548 & \textbf{0.560} \\
& LIQE $\uparrow$ & 2.927 & 2.798 & 3.243 & 4.126 & 3.894 & 3.942 & 3.964 & \textbf{4.150} \\
\hline
\multirow{5}{*}{RealSR} 
& LPIPS $\downarrow$ & 0.271 & 0.316 & 0.306 & 0.299 & 0.347 & 0.292 & 0.319 & \textbf{0.250} \\
& MUSIQ $\uparrow$ & 60.370 & 56.892 & 65.653 & 69.675 & 68.340 & 69.087 & 68.073 & \textbf{70.100} \\
& MANIQA $\uparrow$ & 0.551 & 0.511 & 0.622 & 0.643 & 0.653 & 0.634 & 0.661 & \textbf{0.672} \\
& ClipIQA $\uparrow$ & 0.432 & 0.407 & 0.472 & 0.577 & 0.586 & 0.552 & 0.550 & \textbf{0.590} \\
& LIQE $\uparrow$ & 3.358 & 2.853 & 3.750 & 4.123 & 4.026 & 4.065 & 3.977 & \textbf{4.180} \\
\hline
\multirow{4}{*}{RealLR200} 
& MUSIQ $\uparrow$ & 62.961 & 59.695 & 63.433 & 69.428 & 68.027 & 69.547 & 70.469 & \textbf{70.752} \\
& MANIQA $\uparrow$ & 0.553 & 0.525 & 0.579 & 0.612 & 0.629 & 0.606 & 0.645 & \textbf{0.662} \\
& ClipIQA $\uparrow$ & 0.451 & 0.452 & 0.458 & 0.566 & 0.582 & 0.551 & 0.588 & \textbf{0.598} \\
& LIQE $\uparrow$ & 3.484 & 3.054 & 3.379 & 4.006 & 4.003 & 4.069 & 4.331 & \textbf{4.769} \\
\hline
\multirow{4}{*}{RealLQ250} 
& MUSIQ $\uparrow$ & 62.514 & 59.337 & 56.858 & 70.556 & 69.876 & 69.580 & 71.832 & \textbf{72.322} \\
& MANIQA $\uparrow$ & 0.524 & 0.500 & 0.504 & 0.594 & 0.624 & 0.578 & 0.632 & \textbf{0.663} \\
& ClipIQA $\uparrow$ & 0.435 & 0.417 & 0.382 & 0.562 & 0.578 & 0.528 & 0.578 & \textbf{0.597} \\
& LIQE $\uparrow$ & 3.341 & 2.753 & 2.719 & 4.005 & 4.003 & 3.904 & 4.356 & \textbf{4.412} \\
\hline
\end{tabular}}
\caption{Quantitative comparison of RSR methods on four real-world benchmarks. Our method achieves SOTA performance across four benchmarks.}
\label{tab1}
\end{table}

\begin{figure*}[th]
\centering
\includegraphics[width=2\columnwidth]{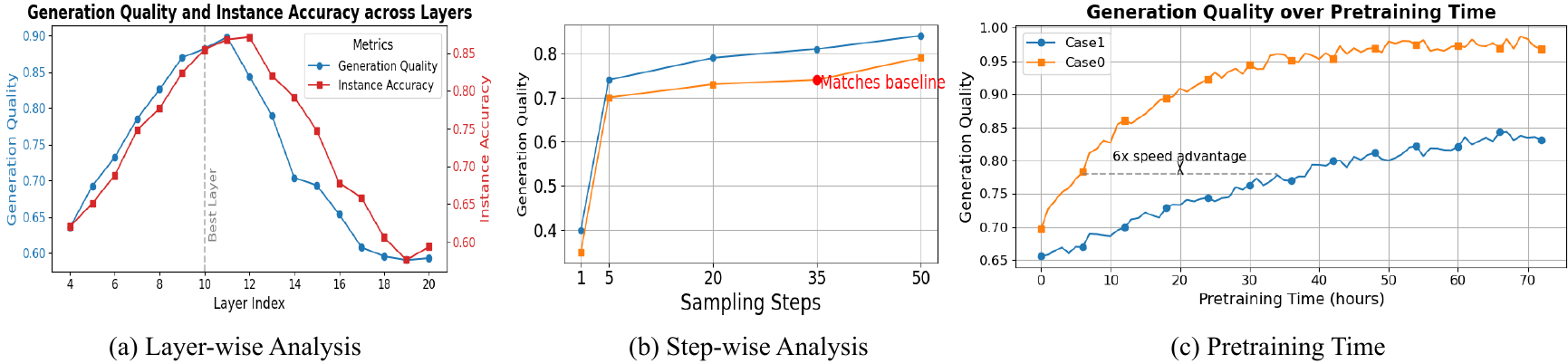} 
\caption{Ablation study analysis. (a) Representation learning trained with intermediate features $\textbf{f}$ extracted from different layers. (b) Effect of varying sampling steps under the default setting. (c) Impact of representation learning on pre-training efficiency and reconstruction quality.}
\label{fig4}
\end{figure*}

\section{Experiments}
\subsection{Experimental Setup}


We pre-train on the Segment Anything dataset comprising 1B images, with all images resized to 512×512. DiT XL/2 serves as the backbone for pretraining. Training is conducted on an H100 GPU with a batch size of 256 for 72 hours, following the default training parameters and strategies. The loss weights of REPA and IS are set to 0.5 and 0.1, respectively. We evaluate extensively on perceptual and quality metrics, including LPIPS \cite{zhang2018unreasonable}, MUSIQ \cite{ke2021musiq}, MANIQA \cite{yang2022maniqa}, CLIPIQA \cite{wang2023exploring}, and NIQE \cite{zhang2023blind}, and benchmark against a wide range of SOTA RSR models, such as Real-ESRGAN \cite{wang2021real}, ResShift \cite{yue2023resshift}, StableSR \cite{wang2024exploiting}, SeeSR \cite{wu2024seesr}, DiffBIR \cite{lin2024diffbir}, OSEDiff \cite{wu2024one}, and DiT4SR \cite{duan2025dit4sr}.

\subsection{Quantitative Evaluation}


We conduct a comprehensive quantitative comparison on four widely adopted real-world benchmark datasets: DrealSR \cite{wei2020component}, RealSR \cite{cai2019toward}, RealLR200 \cite{wu2024seesr}, and RealLQ250 \cite{ai2024dreamclear}. For fair evaluation, we adopt the official results reported by DiT4SR as the baseline for all competing methods, thereby ensuring consistency in the evaluation protocol. Our method generates outputs at a resolution of 512 pixels and subsequently resizes them to the target scale, which is aligned with standard practice.

As shown in Table \ref{tab1}, our approach consistently achieves the best performance across all perceptual and no-reference image quality assessment metrics. In terms of pixel-level consistency, our method yields the lowest LPIPS scores, demonstrating superior preservation of structures and fine details in the reconstructed images. Regarding no-reference metrics such as MUSIQ, MANIQA, and ClipIQA, our method achieves significant improvements, highlighting the role of representation learning in enhancing perceptual naturalness and overall visual quality. Moreover, our approach attains the best performance on the LIQE metric, further confirming its ability to restore artifact-free, natural, and high-quality images.

\subsection{Qualitative Evaluation}

In the visualization results on the RealSR dataset, InstanceRSR demonstrates superior performance in recovering fine-grained instance details compared to representative baselines. Competing approaches often suffer from geometric distortions, over-smoothing, or noise/bleeding artifacts, which blur structural elements such as windows, petals, and poles (highlighted in green/red boxes). In contrast, our method faithfully reconstructs sharp window panes, realistic textures and well-defined boundaries, while maintaining artifact-free depth consistency and structural coherence.

\begin{table}[t]
\centering
\small
\renewcommand{\arraystretch}{1.2}
\setlength{\tabcolsep}{6pt}
\scalebox{0.8}{\begin{tabular}{c|ccc}
\hline
Case & Quality $\uparrow$ & Efficiency $\uparrow$ & Instance Awareness $\uparrow$ \\
\hline
Case 0 (Default) & 1.00 & 1.00 & 1.00 \\
Case 1 (w/o Rep. Learning) & 0.92 & 0.65 & 0.93 \\
Case 2 (w/o Mask Modeling) & 0.94 & 0.90 & 0.85 \\
\hline
\end{tabular}}
\caption{Ablation study results normalized to the default setting (Case 0 = 1).}
\label{tab2}
\end{table}

\subsection{Ablation Study}





We conduct ablation studies to evaluate the impact of instance masks and representation learning on generation quality, pretraining efficiency, and instance awareness. Specifically, Case 1 denotes the model without representation learning, and Case 2 denotes the model without joint modeling of instance masks, both compared against the default Case 0. As shown in Table \ref{tab2}, the default setting achieves the best performance across quality, efficiency, and instance-aware metrics. The comparison of Case 1 with Case 0 and Case 2 shows a significant drop in pretraining speed, demonstrating the effectiveness of representation learning. Meanwhile, Case 2 clearly highlights the substantial improvement in instance awareness brought by joint mask modeling.

\textbf{Layer analysis}: We employ the linear probing technique from REPA to analyze features across different layers to identify the optimal layer. As shown in Fig. \ref{fig4}(a), features from layer 10 yield the best performance in both generation quality and instance accuracy.

\textbf{Sampling step analysis}: Benefiting from representation alignment, when using DDIM for accelerated inference, 5 steps suffice to match the baseline performance, while 10 steps not only surpass the baseline but also significantly outperform it at the same step count. This demonstrates the superior efficiency of InstanceRSR.

\textbf{Pretraining efficiency}: As shown in Fig. \ref{fig4}(c), representation learning enables the model to achieve high-quality outputs early in training, with an approximate 6× speedup. As pretraining continues, its performance consistently surpasses that of Case 1.

\section{Conclusion}


We propose InstanceRSR, a novel real-world super-resolution (RSR) method that integrates instance awareness with representation learning. Our approach employs low-resolution (LR) images as a global conditioning signal to ensure consistent super-resolved outputs, while jointly modeling semantic segmentation to enhance instance-level perception. Furthermore, representation learning is incorporated to guide the model toward improved quality and instance awareness. Innovatively, we introduce scale-supervised instance refinement to further strengthen detail preservation. Extensive experiments on four real-world datasets demonstrate state-of-the-art performance across multiple quantitative metrics, and our method achieves significant improvements in visual quality, particularly in recovering fine-grained instance details.

\bibliographystyle{IEEEbib}
\bibliography{strings,refs}

\end{document}